\documentclass[10pt,twocolumn,letterpaper]{article}

\usepackage{wacv}              

\usepackage{graphicx}
\usepackage{amsmath}
\usepackage{amssymb}
\usepackage{booktabs}

\usepackage{algpseudocode}
\usepackage{algorithm}
\usepackage{algorithmicx}
\usepackage{dsfont}

\usepackage{caption}
\usepackage{amsfonts} 

\usepackage[pagebackref,breaklinks,colorlinks]{hyperref}

\usepackage[capitalize]{cleveref}
\crefname{section}{Sec.}{Secs.}
\Crefname{section}{Section}{Sections}
\Crefname{table}{Table}{Tables}
\crefname{table}{Tab.}{Tabs.}

\def\wacvPaperID{981} 
\def\confName{WACV}
\def\confYear{2025}

\begin{document}

\title{DocTTT: Test-Time Training for Handwritten Document Recognition Using Meta-Auxiliary Learning}

\author{Wenhao Gu, Li Gu, Ziqiang Wang, Ching Yee Suen, Yang Wang\\
Department of Computer Science and Software Engineering, Concordia University\\
{\tt\small \{wenhao.gu,li.gu,ziqiang.wang\}@mail.concordia.ca, \{chingyee.suen, yang.wang\}@concordia.ca}
}
\maketitle

\begin{abstract}
Despite recent significant advancements in Handwritten Document Recognition (HDR), the efficient and accurate recognition of text against complex backgrounds, diverse handwriting styles, and varying document layouts remains a practical challenge. Moreover, this issue is seldom addressed in academic research, particularly in scenarios with minimal annotated data available.
In this paper, we introduce the DocTTT framework to address these challenges. The key innovation of our approach is that it uses test-time training to adapt the model to each specific input during testing. We propose a novel Meta-Auxiliary learning approach that combines Meta-learning and self-supervised Masked Autoencoder~(MAE). During testing, we adapt the visual representation parameters using a self-supervised MAE loss. During training, we learn the model parameters using a meta-learning framework, so that the model parameters are learned to adapt to a new input effectively. Experimental results show that our proposed method significantly outperforms existing state-of-the-art approaches on benchmark datasets.
\end{abstract}

\section{Introduction}
\label{sec:intro}

Handwritten Document Recognition (HDR) is an important technology for interpreting handwritten texts at the page level. As shown in Figure~\ref{inputoutput}, given an input image of a hand-written document, the goal of HDR is to parse the content of the document, e.g. by producing an XML file describing various components (e.g. layout, text tokens) of the page. HDR is essential for digitizing historical documents, automating form processing, and improving manuscript accessibility. Its applications range from enhancing archive access and simplifying administrative tasks to aiding document authentication in legal contexts.
\setcounter{figure}{0}
\begin{figure}
    \centering
    \includegraphics[width=1\linewidth]{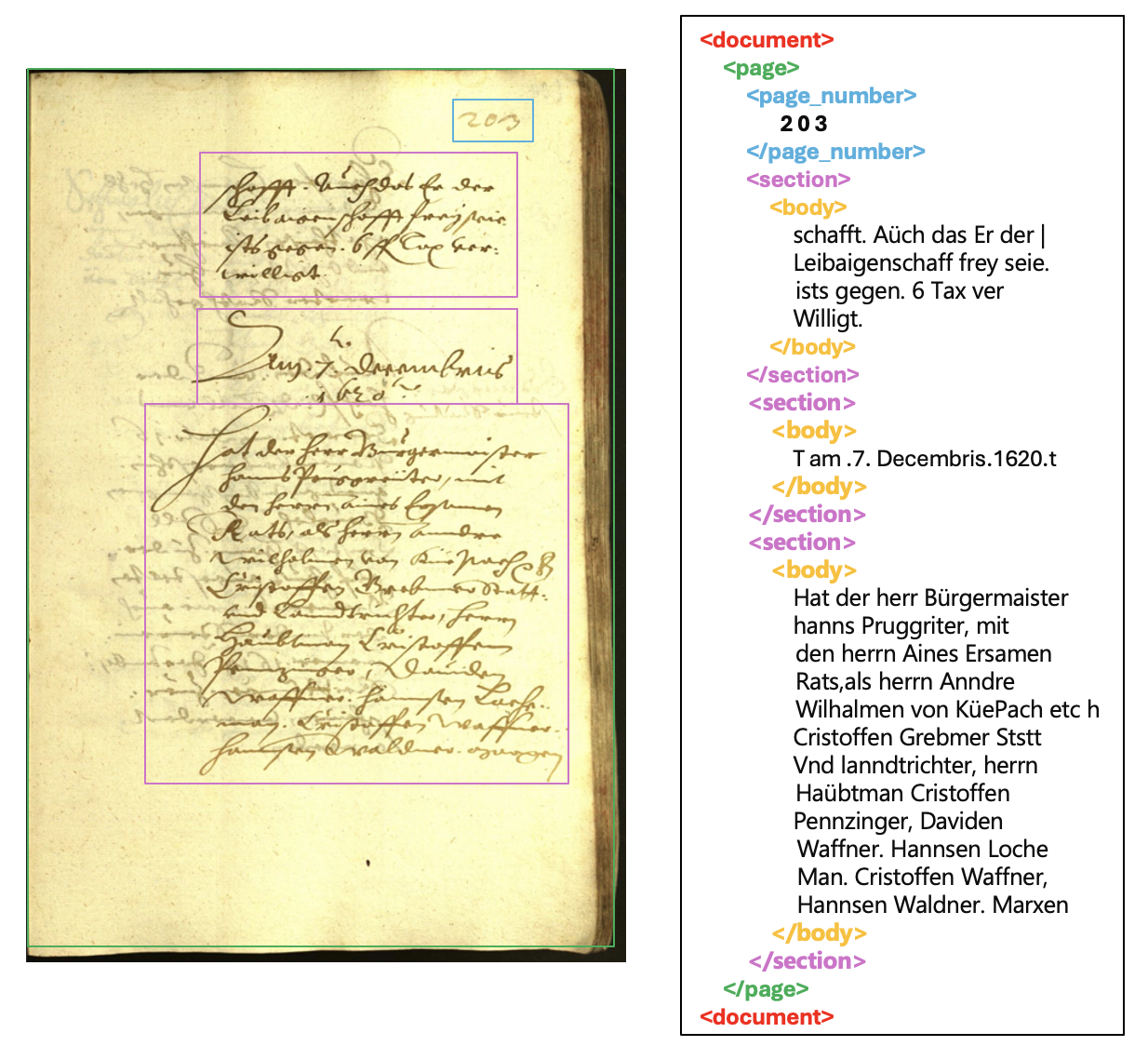}
    \vspace{-20pt}
    \caption{\textbf{Illustration of the handwritten document recognition (HDR) problem.} Given an input image (left) of a handwritten document, the output of HDR is an XML file (right) describing the page layout. The XML-based representation of the page captures the layout and text tokens of the document image~\cite{albahli2021improved,nurseitov2021handwritten,coquenet2023dan}. The XML representation can be flattened as a sequence of tokens. The goal of our work is to predict this sequence of tokens for a given document image.}
    \label{inputoutput}
\end{figure}

Early works~(e.g. \cite{coquenet2022end}) in HDR operate at the line level. These works cannot handle the complexity of the whole page layout. Recent efforts in HDR (e.g. \cite{ghazal2022convolutional, rajalakshmi2019pattern}) propose to understand the whole page of handwritten contents with complex layouts in a supervised manner. These approaches address the limitation of line-level recognition and provide a broader understanding of handwritten documents. Most existing HDR approaches learn a model on a labeled dataset. Then the model is fixed and used for all new inputs during testing. However, due to the unique features associated with the nature of an author's writing style and the method of writing~\cite{hodel2021general}, as shown in Figure~\ref{fig:my4x4grid}, it is difficult for a single model to effectively handle the complexity of all possible documents.
\setcounter{figure}{1}
\begin{figure*}[htbp]
  
  \centering
  \begin{tabular}{cccc}
    \textbf{Ground Truth} & \textbf{Masked Image} & \textbf{Reconstructed A} & \textbf{Reconstructed B} \\
    \begin{subfigure}{0.22\textwidth}
      \includegraphics[width=\linewidth]{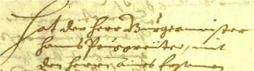}
      
    \end{subfigure} &
    \begin{subfigure}{0.22\textwidth}
      \includegraphics[width=\linewidth]{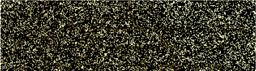}
     
    \end{subfigure} &
    \begin{subfigure}{0.22\textwidth}
      \includegraphics[width=\linewidth]{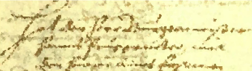}
      
    \end{subfigure} &
    \begin{subfigure}{0.22\textwidth}
      \includegraphics[width=\linewidth]{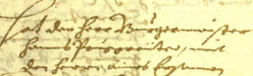}
      
    \end{subfigure} \\
    \begin{subfigure}{0.22\textwidth}
      \includegraphics[width=\linewidth]{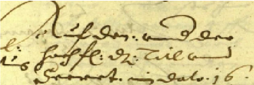}
      
    \end{subfigure} &
    \begin{subfigure}{0.22\textwidth}
      \includegraphics[width=\linewidth]{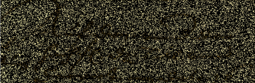}
      
    \end{subfigure} &
    \begin{subfigure}{0.22\textwidth}
      \includegraphics[width=\linewidth]{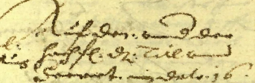}
      
    \end{subfigure} &
    \begin{subfigure}{0.22\textwidth}
      \includegraphics[width=\linewidth]{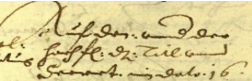}
     
    \end{subfigure} \\
    \begin{subfigure}{0.22\textwidth}
      \includegraphics[width=\linewidth]{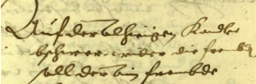}
      
    \end{subfigure} &
    \begin{subfigure}{0.22\textwidth}
      \includegraphics[width=\linewidth]{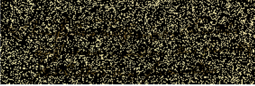}
       
    \end{subfigure} &
    \begin{subfigure}{0.22\textwidth}
      \includegraphics[width=\linewidth]{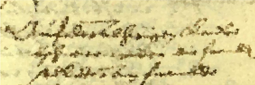}
      
    \end{subfigure} &
    \begin{subfigure}{0.22\textwidth}
      \includegraphics[width=\linewidth]{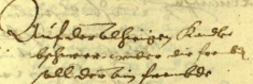}
      
    \end{subfigure} \\
    \begin{subfigure}{0.22\textwidth}
      \includegraphics[width=\linewidth]{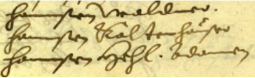}
    \vspace{-25pt}  
    \end{subfigure} &
    \begin{subfigure}{0.22\textwidth}
      \includegraphics[width=\linewidth]{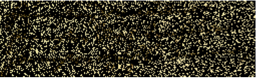}
    \vspace{-25pt}  
    \end{subfigure} &
    \begin{subfigure}{0.22\textwidth}
      \includegraphics[width=\linewidth]{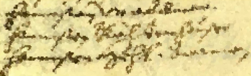}
    \vspace{-25pt}  
    \end{subfigure} &
    \begin{subfigure}{0.22\textwidth}
      \includegraphics[width=\linewidth]{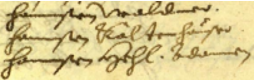}
    \vspace{-25pt}
    \end{subfigure}
  \end{tabular}
  
  \caption{\textbf{Illustration of the capability of Masked Autoencoders.} MAE effectively tackles the HDR problem by reconstructing diverse handwriting styles from unseen test data. The image features four distinct writing styles (first column), a 75\% masked image (second column), and reconstructions from both low-resolution (Reconstructed A) and high-resolution (Reconstructed B) inputs. Accurately predicting token sequences from these varied handwriting styles in document images presents substantial challenges. This illustrates the need for HDR solutions that can adapt to different writing styles.}
  \label{fig:my4x4grid} 
\end{figure*}
In this paper, we propose a new approach that allows the model to adapt to each input during testing. Our work is inspired by the success of test-time training (TTT)~\cite{sun2020test} for image classification, where a self-supervised auxiliary task (e.g. rotation prediction) is used to update the model parameters during inference for a test instance. The updated model is then used to solve the primary task (e.g. image classification). Some recent work~\cite{chi2021test,liu2023meta,wu2023metagcd,wu2024test} shows naively training the primary and auxiliary tasks together may not be optimal, since the model may focus on improving the auxiliary task instead of the primary task. Following \cite{chi2021test}, we propose a new approach (DocTTT) of test-time training for HDR. Our method leverages the model-agnostic meta-learning (MAML)~\cite{finn2017modelagnostic}. We treat each training instance as a task~\cite{gu2022improving,zhong2022meta}. We define a self-supervised MAE loss as the auxiliary branch in the model. For each task, model parameters are updated using the auxiliary task loss, then applied to the primary HDR task. The model is trained using a bi-level optimization Meta-Auxiliary approach, where the auxiliary task explicitly enhances the primary task.

The contributions of our work are manifold. First, we introduce a new test-time adaption approach for handwritten document recognition. By adapting the model during testing, our approach can effectively handle handwritten documents with diverse handwriting styles by quickly capturing the unique visual features of different authors. Second, we integrate Meta-Auxiliary learning with TTT by updating the model with the self-supervised MAE loss associated with the auxiliary task, then applying it to the primary HDR task. Both task parameters are jointly trained through Meta-Auxiliary learning which optimizes the model to enhance the primary task performance using updates from the auxiliary task. Our approach's effectiveness is demonstrated against other leading methods on benchmark datasets.
\setcounter{figure}{2}
\begin{figure*}
    \centering
    \includegraphics[width=0.95\linewidth]{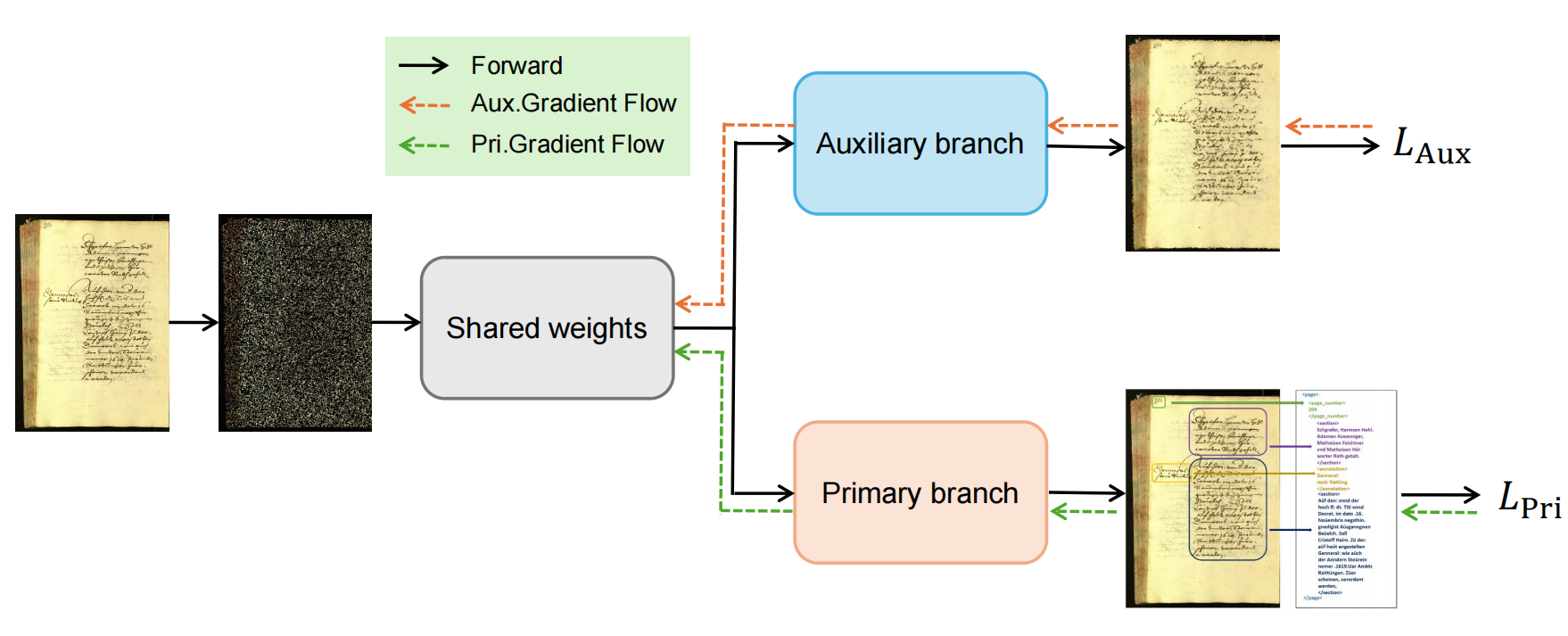}
    \vspace{-15pt}
    \caption{\textbf{Illustration of our model architecture.} On the left, the input document image is masked and then passes through shared weights to the auxiliary branch for the reconstruction task. We use the auxiliary branch to fine-tune the model for each instance using a self-supervised masked autoencoder loss. The adapted model is then used for the primary task of predicting the flattened XML representation (see Figure~\ref{inputoutput}) as a sequence of tokens. }
    \label{Architecture}
\end{figure*}

\section{Related Works}
This section provides an overview of the key areas connected to our study, including handwritten document recognition, meta-learning, and mask autoencoders.

\subsection{Handwritten Document Recognition}
Advancements in handwritten document recognition (HDR) have been driven by deep learning approaches like CNNs \cite{ahlawat2020improved}, which extract features directly from images. The integration of RNNs, particularly LSTMs, has improved the modeling of handwriting's sequential nature \cite{Johnson2021,dutta2018improving}. More recently, transformer models have offered new capabilities in handling complex handwriting patterns without needing pre-segmentation, thanks to their self-attention mechanisms \cite{Anderson2022}. The transition to end-to-end models facilitates the direct transcription of handwritten texts, supported by extensive datasets and benchmarks that enhance model training and evaluation \cite{Williams2023}.

\subsection{Meta-Learning}
Meta-learning optimizes learning from minimal data, essential in tasks like few-shot learning~\cite{liu2022few,chi2022metafscil,ma2024crowd} and rapid adaptation to new tasks~\cite{chi2024adapting}. Techniques such as MAML \cite{finn2017modelagnostic} and its variants like Reptile \cite{nichol2018reptile} and Meta-SGD \cite{li2017metasgd} have shown significant utility. Specifically, meta-learning has been applied to adapt quickly to different handwriting styles in character recognition tasks \cite{bartler2022mt3, otsuzuki2021meta, qiu2021meta}. Our work extends these applications to full document recognition, considering both layout and content.

\subsection{Masked Autoencoders}
Masked Autoencoders (MAEs), used in self-supervised learning, reconstruct masked parts of data, such as text or image patches, to learn robust data representations \cite{he2022masked, devlin2018bert, yin2022vit,liang2022self}. This approach is particularly effective in handwriting recognition, allowing models to adapt to individual writing styles by focusing on the unique aspects of masked handwriting samples \cite{lyu2022maskocr}.

\subsection{Test-time Training}
Test-time Training (TTT) enhances model performance on new, unseen data by adjusting pre-trained models during inference to better match the test data's distribution \cite{chen2023improved,wang2025distribution}. Techniques like Tent, which minimizes prediction entropy \cite{wang2020tent}, and self-supervised approaches using pseudo-labels \cite{chen2022contrastive} have proven effective. TTT is especially beneficial in handwriting recognition, adapting models to varied writing styles, ensuring accuracy across diverse datasets.
\begin{figure*}
    \centering
    \includegraphics[width=1\linewidth]{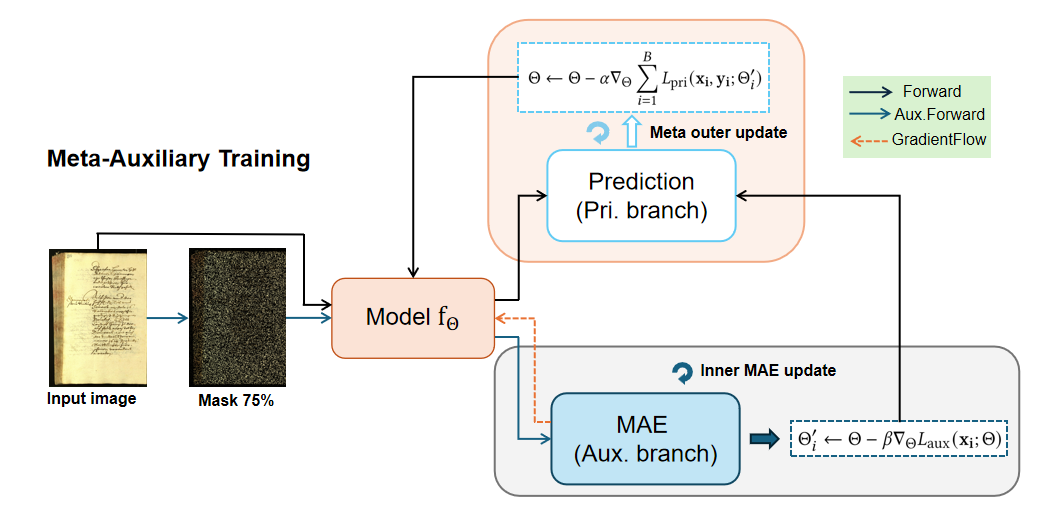}
    \vspace{-28pt}
    \caption{\textbf{Illustration of the Meta-Auxiliary training.} For a training example $\mathbf{x}_i$ with the ground-truth label $\mathbf{y}_i$, we update the model parameters $\Theta$ using a self-supervised MAE loss (inner loop) to obtain an adapted model $\Theta'$ for this training example. We then use the adapted model $\Theta'$ for the primary task. We use the supervised loss for the primary task as the meta-objective in the outer loop to update the model parameter $\Theta$.}
    \label{Meta-contrastive training}
\end{figure*}

\section{Proposed Approach}
\setcounter{figure}{3}

Given an input image of a hand-written document, the goal of HDR is to parse the layout and the content within the image. Let us denote the input image as \(X\) with an expected output as a sequence of tokens \(Y\) with length \(L_Y\).  Following \cite{coquenet2023dan}, we represent the output $Y$ as a sequence of tokens from a dictionary \(D = \text{\textless sos\textgreater}\cup A \cup S \cup \text{\textless eot\textgreater}\) consisting of alphabetic characters \(A\), layout markers \(S\), and special start and an end-of-transcription markers \{\text{\textless sos\textgreater},  \(\text{\textless eot\textgreater}\)\}. This sequence of tokens can be interpreted as the flattened XML file (see Figure~\ref{inputoutput}) representing the layout and content of the document image. In the following, we first describe the model architecture~(Sec.~\ref{sec:architecture}), then introduce the Meta-Auxiliary training framework for learning the model parameters (Sec.~\ref{sec:metatrain}). We also introduce a two-phase training strategy (Sec.~\ref{sec:details}) to speed up the learning process. 

\subsection{Model Architecture}\label{sec:architecture}
Following \cite{chi2021test}, our model architecture consists of two branches. The first branch is for solving the primary task of HDR. The output of this branch is the sequence of tokens. We add another auxiliary branch. This auxiliary branch solves a self-supervised task (e.g. masked autoencoder in this paper). These two tasks share the backbone for extracting image features. During testing, the auxiliary task is used to update the parameters of the shared backbone. Figure~\ref{Architecture} shows the details of the model architecture.

\noindent\textbf{Primary Task.} In the DocTTT framework, the main task is to process a document image $X$, then produce a sequence of tokens \(Y\) representing the layout and content of the document. We use an encoder-decoder architecture. The encoder is a Fully Convolutional Network (FCN) that transforms an input document image \(X\) into a two-dimensional feature map. We add positional information to the 2D feature map and convert it into a one-dimensional feature sequence as the input to the decoder. A transformer-based NLP module is used as the decoder\cite{vaswani2017attention,coquenet2023dan}. The decoder translates the 1D feature sequence produced by the encoder into a series of tokens representing the text and layout of the document. This decoder starts with a start-of-transcription token and sequentially adds predicted tokens, i.e. \(Y=(\hat{y}_0, \hat{y}_1, \ldots, \hat{y}_{t-1})\), until it predicts the end-of-transcription token \(\langle\text{eot}\rangle\).

\noindent\textbf{Self-supervised Auxiliary Task.} In addition to the primary task, our model architecture also has an auxiliary branch corresponding to a self-supervised auxiliary task. A properly chosen auxiliary task can act as a regularization and improve the primary task. The loss function of this auxiliary task will be used for test-time training. The auxiliary task branch shares the backbone feature extractor with the primary task branch.

\setcounter{figure}{4}
\begin{figure}
    \centering
    \includegraphics[width=1\linewidth]{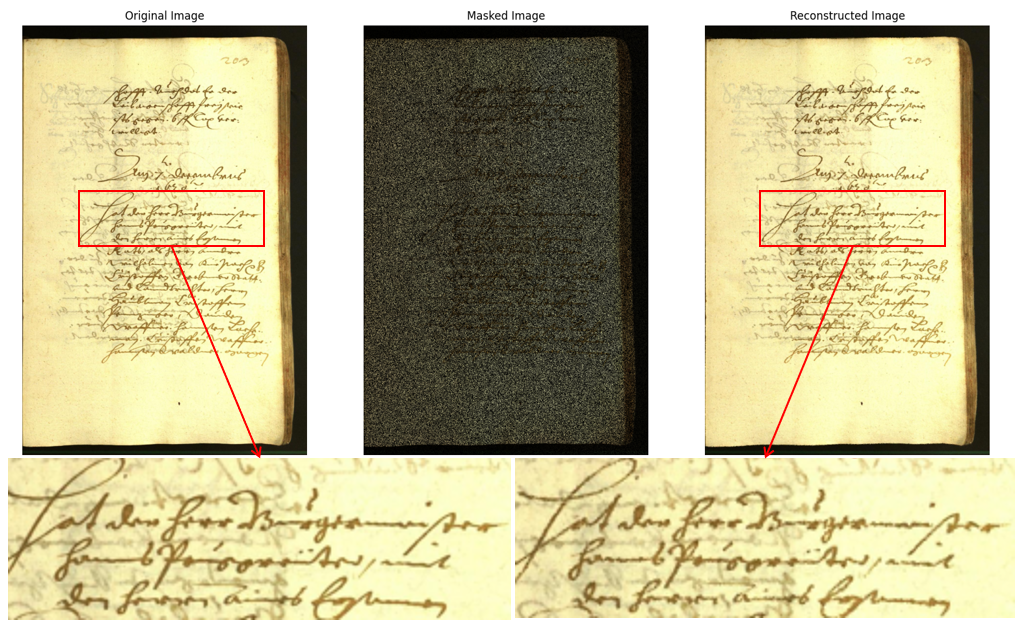}
    \vspace{-20pt}
   \caption{Illustration of the auxiliary branch. Given an input image (top left), we first generate a masked version (top middle) of the input image. The output of the auxiliary branch outputs a reconstruction (top right) of the input image based on its masked version. The bottom two images show the zoomed-in regions of the input and the reconstruction images.}
   \label{contrastive learning}
\end{figure}

In our work, we choose to use Masked Autoencoder~\cite{he2022masked} as the auxiliary task. As depicted in Figure \ref{contrastive learning},
it is designed to reconstruct an input image from a partially masked version. The masking mechanism randomly hides parts of the input image, e.g. by setting a certain percent of the pixel values to zero. This creates a ``masked'' version of the image. The autoencoder's task is to predict the missing parts based solely on the visible pixels.

Our masked autoencoder is divided into an encoder and a decoder. The encoder processes the partially visible input, extracting and compressing essential features into a condensed representation. The decoder then uses this representation to reconstruct the original image, filling in the masked portions by generating the missing details based on learned contextual cues. This process not only pre-trains the CNN layers on unlabeled data but also aids the primary task branch by improving its generalization capabilities. 

\subsection{Meta-Auxiliary Training}\label{sec:metatrain}

\begin{algorithm}
\caption{Meta-Auxiliary Training}
\label{alg:meta_contrastive} 
\begin{algorithmic}[1]
\Require Learning rate $\alpha$ for the outer loop and $\beta$ for the inner loop. 
\State Initial pre-trained weights: $\Theta=\{\theta^S, \theta^{P}, \theta^{A}\}$
\State training example $(\mathbf{x}_i, \mathbf{y}_i)$ for each image
\While{not converged}
    
        \For{number of iterations of meta inner loop}
            \State $\Theta'_i \leftarrow \Theta - \beta \nabla_{\Theta} L_{\text{aux}}(\mathbf{x}_i;\Theta )$
            
        \EndFor

    \State Update global meta-parameters with gradient descent:
    \State $\Theta \leftarrow \Theta - \alpha \nabla_{\Theta} \sum_{i=1}^B L_{\text{pri}}(\mathbf{x}_i,\mathbf{y}_i;\Theta'_i)$
\EndWhile
\State \Return optimized meta-parameters $\Theta$

\end{algorithmic}
\end{algorithm}

We use $\Theta=\{\theta^S, \theta^{P}, \theta^{A}\}$ to denote the model parameters, where $\theta^S$, $\theta^{P}$, $\theta^{A}$ correspond to the parameters of the shared backbone, the primary branch, and the auxiliary branch in the model architecture, respectively. Our Meta-Auxiliary training is based on MAML~\cite{finn2017modelagnostic} which involves a bi-level optimization procedure, which shown in Algorithm \ref{alg:meta_contrastive} and Figure \ref{Meta-contrastive training}.

Given a training examples $\{\mathbf{x}_i,\mathbf{y}_i\}$
 where $\mathbf{x_i}$ is the input image of the $i$-th example and $\mathbf{y}_i$ is the corresponding ground-truth label, we mask $\mathbf{x}_i$ to generate a masked image. We then perform adaptation on each masked example by taking a few gradient updates according to the loss of the auxiliary task as follows:  
\begin{equation}
\Theta'_i \leftarrow \Theta - \beta \nabla_{\Theta} L_{\text{aux}}(\mathbf{x}_i;\Theta )
\label{eq:inner}
\end{equation}
where $L_{\text{aux}}$ is the self-supervised loss of the auxiliary branch and $\beta$ is the learning rate. Note that since the parameters $\theta^{P}$ is not involving in $L_{\text{aux}}$, this update will not change $\theta^{P}$, i.e. ${\theta'}_i^{P}=\theta^{P}$.

We choose the Structural Similarity Index Measure (SSIM) over Mean Squared Error (MSE) for our loss function in image reconstruction tasks involving handwritten content. These tasks are highly sensitive to image resizing and resolution changes seen as in Figure~\ref{fig:my4x4grid}. SSIM is preferable as it evaluates similarity based on texture, luminance, and key for preserving the perceptual quality of the reconstructed images\cite{sara2019image}. This is critical in handwritten text recognition (HTR), where maintaining the detailed features of the original image is essential for accurate text recognition. Unlike natural images, where aesthetic quality might be the focus, HTR requires that every subtle feature of the handwriting, including minute protrusions and indentations, is captured accurately to ensure reliable algorithmic interpretation. This approach is vital in applications like image super-resolution and photo restoration, where preserving original details is crucial\cite{setiadi2021psnr}. Given two images $x$ and $y$, the loss is defined as
\begin{equation}
L_{\text{aux}} = 1-\text{SSIM}(x, y) = 1-\frac{(2\mu_x \mu_y + a)(2\sigma_{xy} + b)}{(\mu_x^2 + \mu_y^2 + a)(\sigma_x^2 + \sigma_y^2 + b)}
\end{equation}
where $\mu_x$ and $\mu_y$ denote the mean intensity values of images $x$ and $y$, respectively. The terms $\sigma_x^2$ and $\sigma_y^2$ represent the variances of $x$ and $y$, while $\sigma_{xy}$ indicates the covariance between the two images. The constants $a$ and $b$ are used to stabilize the division when the denominator is small, ensuring the robustness of the SSIM calculation. Compared with MSE, SSIM is more robust against potential size alternations during the reconstruction process.


Note that since $L_{\text{aux}}$ does not require ground-truth annotation, the update in Eq.~\ref{eq:inner} can be performed at testing. The updated model $\Theta'_i$ can be interpreted as a model adapted to $\mathbf{x_i}$. We use $\Theta'_i$ to perform the primary task. Ideally, we would like the output of $\Theta'_i$ for the primary task to be close to the ground-truth $\mathbf{y_i}$. This can be achieved by optimizing the following meta-objective as follows:
\begin{equation}
\Theta \leftarrow \Theta - \alpha \nabla_{\Theta} \sum_{i=1}^B L_{\text{pri}}(\mathbf{x}_i,\mathbf{y}_i;\Theta'_i)
\label{eq:outer}
\end{equation}
where $\alpha$ the meta-learning rate. The index \( i \) iterates over the batch of data points and \( L_{pri} \) is a supervised loss between the predicted and ground-truth outputs in the primary task. Note that $L_{pri}(\cdot)$ in Eq.~\ref{eq:outer} is a function of the adapted model parameters $\Theta'_i$, but the optimization is over the original model parameters $\Theta$.

For the decoder at each timestep \(t\), the model receives a one-dimensional vector of visual features along with the series of previously predicted tokens, \(\hat{y}_0, \ldots, \hat{y}_{t-1}\), as its inputs. It then calculates the probability distribution \(\mathbf{p_t}\) across all tokens in the dictionary \(D\) for the current timestep\cite{coquenet2023dan}. The cross-entropy loss $L_{CE}$ is used to optimize the parameters of the model by comparing the predicted probability distribution on $\mathbf{p_t}$ against the ground-truth token $\hat{y_t}$: 
\begin{equation}
L_{\text{pri}} = \sum_{t=1}^{L_y+1} L_{CE}(\mathbf{p_t}, \hat{y_t})
\end{equation}

The function $L_{CE}(\mathbf{p_t},\hat{y_t})$ computes the loss for each instance by taking the negative log-likelihood of the true class's predicted probability in the primary task.
 $L_{\text{pri}}$ measures the performance of the adapted model parameters $\Theta'_i$. The meta-learning methodology iteratively refines the global model parameters $\Theta$ in such a way that the learned model can be effectively adapted to a new task via the self-supervised MAE loss.


\noindent {\bf Testing.} During testing, we refine the meta-learned parameters $\Theta$ for a specific test sample \(\mathbf{x}_{\text{test}}\) by generating masked \(\mathbf{x}_{\text{Mtest}}\) and computing SSIM loss. We apply the same loss methodology as in the meta-training phase for the Masked Autoencoders (MAEs) image reconstruction. We then update the model parameters $\Theta$ by taking a few gradient updates to minimize the loss \(L_{\text{aux}}\) in Eq.~\ref{eq:inner} to obtain the updated model parameters $\Theta'$. We then use $\Theta'$ to predict the output of the primary task for $\mathbf{x}_{\text{test}}$.

\subsection{Two-Phase Training}\label{sec:details}
To speed-up the learning process, we implement a two-phase training strategy.

\noindent\textbf{Phase 1: Pre-training.}  In the pre-training phase, our model undergoes pre-training on synthetic printed lines to acquire feature extraction capabilities through a line-level OCR model~\cite{goel2023handwritten}.
The model processes mini-batches of 16 and employs a combination of preprocessing, data augmentation, and curriculum dropout techniques~\cite{coquenet2023dan}. Training begins with simple documents, progressively adding complexity through synthetic data~\cite{biswas2021docsynth}. Dynamic augmentations like resolution changes and color adjustments are applied to synthetic and real images for robustness~\cite{coquenet2021span}. Additionally, teacher forcing with a calibrated error rate and a tailored dropout schedule are used to handle real-world inaccuracies and prevent overfitting, enhancing learning across diverse document styles~\cite{singh2021full}.



\noindent\textbf{Phase 2: Meta-Training.} During the meta-training phase, our model employs a curriculum learning strategy, initially focusing on 90\% synthetic articles and gradually incorporating more real samples, reducing synthetic data to 20\% over time~\cite{rouhou2022transformer,coquenet2022end}. This keeps the synthetic documents to simulate new training scenarios while enabling fine-tuning with real-world samples~\cite{coquenet2023dan}. Techniques like data augmentation enhance resilience to input variations, and teacher forcing introduces controlled errors to prepare for prediction inaccuracies~\cite{coquenet2021span}. The training is complemented with post-processing to preserve the document's structural and grammatical integrity~\cite{coquenet2023dan}.


 Our model is trained on a single NVIDIA RTX 4090 GPU. We use the Adam optimizer starting with a learning rate of $10^{-4}$. The same set of hyperparameters are used across all datasets.

\begin{table}[ht]
\centering

\setlength{\tabcolsep}{3pt} 
\begin{scriptsize}
\begin{tabular}{lcccc}
\hline
\textbf{Method} & \textbf{CER (\%)}$\downarrow$ & \textbf{WER (\%)}$\downarrow$ & \textbf{LOER (\%)}$\downarrow$ & \textbf{mAPCER}$\uparrow$ \\
\hline
SPAN~\cite{coquenet2021span} & 6.20 & 25.69 & -- & -- \\
DefConv 1D-LSTM~\cite{cascianelli2022boosting} & 5.2 & 23.7 & -- & -- \\
DefConv CRNN~\cite{cascianelli2022boosting} & 4.5 & 21.7 & -- & -- \\
DAN~\cite{coquenet2023dan} & 3.43 & 13.05 & 5.17 & 93.32 \\
Faster DAN~\cite{coquenet2023faster} & 3.95 & 14.06 & 3.82 & 94.20 \\
Ours (DocTTT) & \textbf{3.18} & \textbf{12.63} & \textbf{3.71} & \textbf{94.56} \\
\hline
\end{tabular}
\end{scriptsize}
\caption{\textbf{Evaluation of DocTTT on the READ 2016 test set at page level.} For mAPCERE, large numbers mean better performance. For other metrics, small numbers mean better performance.}
\label{tab:read_2016_single_page_level}
\end{table}

\begin{table}[ht]
\centering

\begin{scriptsize}
\begin{tabular}{lcccc}
\hline
\textbf{Method} & \textbf{CER}$\downarrow$ & \textbf{WER}$\downarrow$ & \textbf{LOER}$\downarrow$ & \textbf{mAPCER}$\uparrow$ \\
\hline
DAN~\cite{coquenet2023dan} & 3.70\% & 14.15\% & 4.98\% & 93.09\% \\
Faster DAN~\cite{coquenet2023faster} & 3.88\% & 14.97\% & \textbf{3.08}\% & 94.54\% \\
Ours (DocTTT) & \textbf{3.47}\% & \textbf{13.82}\% & 3.35\% & \textbf{95.18}\% \\
\hline
\end{tabular}
\end{scriptsize}
\caption{\textbf{Evalutation of DocTTT using the READ 2016\cite{sanchez2016icfhr2016} test set at double page level.} For mAPCERE, large numbers mean better performance. For other metrics, small numbers mean better performance.}
\label{tab:read_2016_double_page_level}
\end{table}

\begin{table}[ht]
\centering

\begin{tabular}{lcc}
\hline
\textbf{Method} & \textbf{CER}$\downarrow$ & \textbf{WER}$\downarrow$ \\
\hline
Easter2.0~\cite{chaudhary2022easter2} & 6.21\% & --  \\
Transformer w/ CNN~\cite{kang2005pay}  & 4.67 \% & -- \\
Decouple Attention Network~\cite{wang2020decoupled} & 6.4\% & 19.6\%  \\
SFR~\cite{wigington2018start} & 6.4\% & 23.2\%  \\
SPAN~\cite{coquenet2021span}  & 5.45 \% & 19.83\% \\
VAN~\cite{coquenet2022end} & 4.45\% & 14.55\%  \\
Ours (DocTTT) & \textbf{4.22}\% & \textbf{14.17}\%  \\
\hline
\end{tabular}
\caption{\textbf{Evaluation of DocTTT using the IAM\cite{marti2002iam} test set at paragraph level.} Comparison with state-of-the-art methods.}
\label{tab:IAM}
\end{table}

\begin{table}[ht]
\centering

\begin{tabular}{lcc}
\hline
\textbf{Method} & \textbf{CER}$\downarrow$ & \textbf{WER}$\downarrow$ \\
\hline
SPAN~\cite{coquenet2021span}  & 3.81 \% & 13.8\% \\
Puigcerver et al.~\cite{puigcerver2017multidimensional}*  & 3.3 \% & 12.8\% \\
Coquenet et al.~\cite{coquenet2022end} & 3.04\% & 8.32\%  \\
DAN~\cite{coquenet2023dan} & 2.63\% & 6.78\%  \\
Ours (DocTTT) & \textbf{2.33}\% & \textbf{6.47}\%  \\
\hline
\end{tabular}
\caption{\textbf{Evaluation of DocTTT using the Rimes 2011~\cite{grosicki2011icdar} test set at line level.} Comparison with state-of-the-art methods. *The work in \cite{puigcerver2017multidimensional} uses some postprocessing to further improve the result. For comparison, we report the result without this postprocessing for \cite{puigcerver2017multidimensional}.}
\label{tab:rimes_2011_line_level}
\end{table}

\begin{table*}[ht]
\centering

\resizebox{\textwidth}{!}{%
\begin{tabular}{l|cc|cc|cccc|cccc}
\hline
\textbf{} & \multicolumn{2}{c|}{\textbf{RIMES 2011\cite{grosicki2011icdar} (line)}} & \multicolumn{2}{c|}{\textbf{IAM \cite{marti2002iam} (paragraph)}} & \multicolumn{4}{c|}{\textbf{READ 2016\cite{sanchez2016icfhr2016} (single-page)}} & \multicolumn{4}{c}{\textbf{READ 2016\cite{sanchez2016icfhr2016} (double-page)}} \\
\cline{2-11}
& \textbf{CER}$\downarrow$ & \textbf{WER}$\downarrow$ & \textbf{CER}$\downarrow$ & \textbf{WER}$\downarrow$ &\textbf{CER}$\downarrow$ & \textbf{WER} & \textbf{LOER}$\downarrow$ & \textbf{mAPCER}$\uparrow$ & \textbf{CER}$\downarrow$ & \textbf{WER}$\downarrow$ & \textbf{LOER}$\downarrow$ & \textbf{mAPCER}$\uparrow$ \\
\hline

 Baseline & 2.67 & 6.95 & 4.55 & 14.70 & 3.49 & 13.75 & 4.95 & 93.47 & 3.71 &14.50 & 3.23 & 95.19\\
 DocTTT w/o positional encoding & 76.39 & 79.21 & 100 & $>100$ & 85.66 & 88.08 & 13.97 & 2.74 & 86.22 & 87.53 & 15.68 & 8.26 \\
 DocTTT w/o teacher forcing & 6.73 & 11.30 & 7.54 & 16.88 & 6.52 & 17.23 & 3.56 & 90.87 & 5.73 & 20.19 & 6.75 & 89.42 \\
 DocTTT w/o curriculum dropout & 2.42 &6.53 & 4.69 & 14.78 & 3.77 & 13.68 & 4.50 & 93.39 & 3.87 & 14.52 & 4.38 & 93.76 \\
DocTTT w/o TTT & 2.58 & 6.91 & 4.68 & 14.73 & 3.48 & 13.32 & 4.25 & 94.18 & 3.55 & 13.97 & 3.46 & 94.93 \\
DocTTT w/o meta-learning & 2.46 & 6.97 & 4.71 & 15.08  & 3.31 & 13.61 & \textbf{3.49} & 94.26 & 3.68 &14.59 & 3.27 & \textbf{95.33}\\
 DocTTT (Ours) & \textbf{2.33} & \textbf{6.47} & \textbf{4.22} & \textbf{14.17}& \textbf{3.18} & \textbf{12.63} & 3.71 & \textbf{94.56} & \textbf{3.47} &\textbf{13.82} & \textbf{3.35} & 95.18 \\
\hline
\end{tabular}
}
\caption{\textbf{Ablation study of on the test set of the RIMES 2011\cite{grosicki2011icdar} IAM\cite{marti2002iam}, and READ 2016\cite{sanchez2016icfhr2016} datasets.} We remove various components of our DocTTT framework and report their performance metrics. For mAPCERE, large numbers mean better performance. For other metrics, small numbers mean better performance. Our DocTTT framework achieves the best accuracy in most cases. This demonstrates that each component has a positive impact on the overall performance.}
\label{tab:faster_dan_ablation_study}
\end{table*}



    

\section{Experiments}
\subsection{Datasets}
We evaluate the performance of our method using three benchmark datasets: RIMES2011~\cite{grosicki2011icdar}, READ2016~\cite{sanchez2016icfhr2016}, and the IAM Handwriting Database~\cite{marti2002iam}. These datasets are widely used in the field of handwritten document recognition.

The RIMES2011 dataset, derived from the RIMES project~\cite{grosicki2011icdar}, features over 12,000 pages of handwritten letters from more than 1,300 volunteers, offering diverse writing styles for line-level benchmarking.

The READ2016 dataset~\cite{sanchez2016icfhr2016} includes over 30,000 pages of early modern German scripts from 1470 to 1805, annotated at the line level. It poses a unique challenge with its complex handwriting, used in the ICFHR 2016 HTR competition.

The IAM Handwriting Database contains unconstrained handwritten English text, scanned at 300dpi~\cite{marti2002iam}. Initially introduced at ICDAR 1999, it supports the development of HMM-based recognition systems and writer identification.





\subsection{Metrics}

In this study, we evaluate document recognition by assessing text and layout separately and together. Text recognition is measured using Character Error Rate (CER) and Word Error Rate (WER). These metrics compare predictions to corrected ground truths, excluding layout markers. The CER metric is defined as:
\begin{equation}
    CER = \frac{\sum_{i=1}^{K} d_{\text{lev}}(\hat{y}_i^{\text{text}}, y_i^{\text{text}})}{\sum_{i=1}^{K} \text{y}_{len_i}^{\text{text}}}
\end{equation}
where \(y_{\text{text}}\) is the ground-truth and \(\hat{y}_{\text{text}}\) is the prediction. The difference between them is calculated as the total of the Levenshtein distances (abbreviated \(d_{\text{lev}}\)) between them normalized by the total length of the ground truths represented as \(\text{y}_{len_i}^{\text{text}}\) for K examples. WER is calculated similarly to CER but focuses on words, treating punctuation as separate words~\cite{coquenet2023dan}. 

For layout recognition, we employ the Layout Ordering Error Rate (LOER), which uses a normalized Graph Edit Distance (GED) to accurately measure the structure and relationships within the document~\cite{wang2021combinatorial}. It is defined as:
\begin{equation}
\text{LOER} = \frac{\sum_{i=1}^{K} \text{GED}(y_i^{\text{graph}}, \hat{y}_i^{\text{graph}})}{\sum_{i=1}^{K} (n_{\text{e}_i} + n_{\text{n}_i})} 
\end{equation}
where \(y_i^{\text{graph}}\) is the graph representation of ground truth and \(\hat{y}_i^{\text{graph}}\) is the graph representation of the prediction. The metric is normalized by the sum of the number of edges \(n_{\text{e}_i}\) and nodes \(n_{\text{n}_i}\) for $K$ samples in the dataset~\cite{coquenet2023dan}.

The Mean Average Precision Character Error Rate (mAPCER) evaluates both text and layout recognition by applying object detection principles~\cite{lin2014microsoft}\cite{everingham2010pascal}. This metric uses various CER thresholds to determine text classification accuracy within specific layout areas~\cite{coquenet2023dan}.
\begin{equation}
    mAP_{CER} = \frac{\sum_{c \in S} AP_{\text{CER}_c}^{5:50:5} \cdot len_c}{\sum_{c \in S} len_c}
\end{equation}
where \(AP_{\text{CER}_c}^{5:50:5}\) denotes the mean of the precision values computed by considering several Character Error Rate (CER) thresholds, which range from a minimum of \( \theta_{\min} = 5\% \) to a maximum of \( \theta_{\max} = 50\% \), in increments of \( \Delta\theta = 5\% \). The metric is weighted by the number of characters \(\text{len}_c\) in each class c\cite{coquenet2023dan}.

These metrics offer comprehensive evaluations of the model performance in the nuanced task of jointly identifying and categorizing text and layout elements.

\subsection{Experimental Results}

Apart from DAN~\cite{coquenet2023dan} and Faster DAN~\cite{coquenet2023faster}, no other studies have evaluated systems at the page and double-page levels on the READ 2016 dataset~\cite{sanchez2016icfhr2016} under similar conditions without external data or language models. Our method, DocTTT, is tested on the READ 2016, IAM~\cite{marti2002iam}, and Rimes 2011~\cite{grosicki2011icdar} datasets. We assess performance from single and double pages to paragraphs and lines to ensure our method is effective across various formats. The results are illustrated in Figures \ref{qualitative examples1} and \ref{qualitative examples2}.


\noindent{\bf READ 2016~\cite{sanchez2016icfhr2016} Single Page Level.} Table \ref{tab:read_2016_single_page_level} demonstrates DocTTT's superior performance on the READ 2016 dataset at the single page level, with a CER of 3.18\%, a WER of 12.63\%, a LOER of 3.71\%, and an mAPCER of 94.56\%. This highlights its effectiveness in text recognition and layout understanding.

\noindent{\bf READ 2016~\cite{sanchez2016icfhr2016} Double Page Level.} Table \ref{tab:read_2016_double_page_level} reveals that DocTTT excels in more complex double page scenarios on the READ 2016 dataset, achieving a CER of 3.47\%, a WER of 13.82\%, and mAPCER of 95.18\%. These results demonstrate its robustness and high accuracy in managing intricate document layouts.

\noindent{\bf IAM~\cite{marti2002iam} Paragraph Level.} Table \ref{tab:IAM} shows the results of the paragraph level on the IAM dataset. For our pretraining at the line level, using the same split for training and validation was not feasible because some lines are extracted from the same paragraph image. Therefore, we opted for a similar split to ensure a fair comparison. DocTTT method outperforms all other methods listed, achieving the lowest Character Error Rate (CER) of 4.22\% and a competitive Word Error Rate (WER) of 14.17\%. 

\noindent{\bf Rimes 2011~\cite{grosicki2011icdar} Line Level.} Table \ref{tab:rimes_2011_line_level} shows the results of the line level on the Rimes 2011 dataset. DocTTT once again showcases its strong performance with a CER of 2.33\% and the lowest WER at 6.47\%. The results demonstrate the applicability of DocTTT in scenarios requiring fine-grained text recognition.
Note that the result in \cite{puigcerver2017multidimensional} uses a  4-grams model to help estimate the probability distribution of sequences of characters as a postprocessing. But other methods (including ours) in the table do not use such postprocessing.

\noindent{\bf Discussions.} The consistent outperformance of DocTTT across these evaluations, when compared to methods like DAN, Faster DAN, and research by Coquenet et al. and Puigcerver et al., showcases its comprehensive ability to address the dual challenges of text and layout recognition in document analysis. Notably, DocTTT's advancements are not limited to either text or layout individually but extend to their integration, as evidenced by its superior performance in both CER and WER (for text accuracy) and LOER and mAPCER (for layout and integrated text-layout understanding)\cite{coquenet2023dan}.

We have shown that the DocTTT is robust on a variety of datasets. We have employed the same hyperparameters despite notable variations in layout, language, color encoding, and the quantity of training examples between three different datasets\cite{coquenet2023dan}. We demonstrate that our model can quickly adapt to the new writing style introduced by each image as shown in Figure ~\ref{qualitative examples2}. By utilizing this unsupervised inner loss, we avoid the need for additional annotations. We show that this approach remains highly effective in both complex handwritten datasets.

\setcounter{figure}{5}
\begin{figure}
    \centering
    \includegraphics[width=1\linewidth]{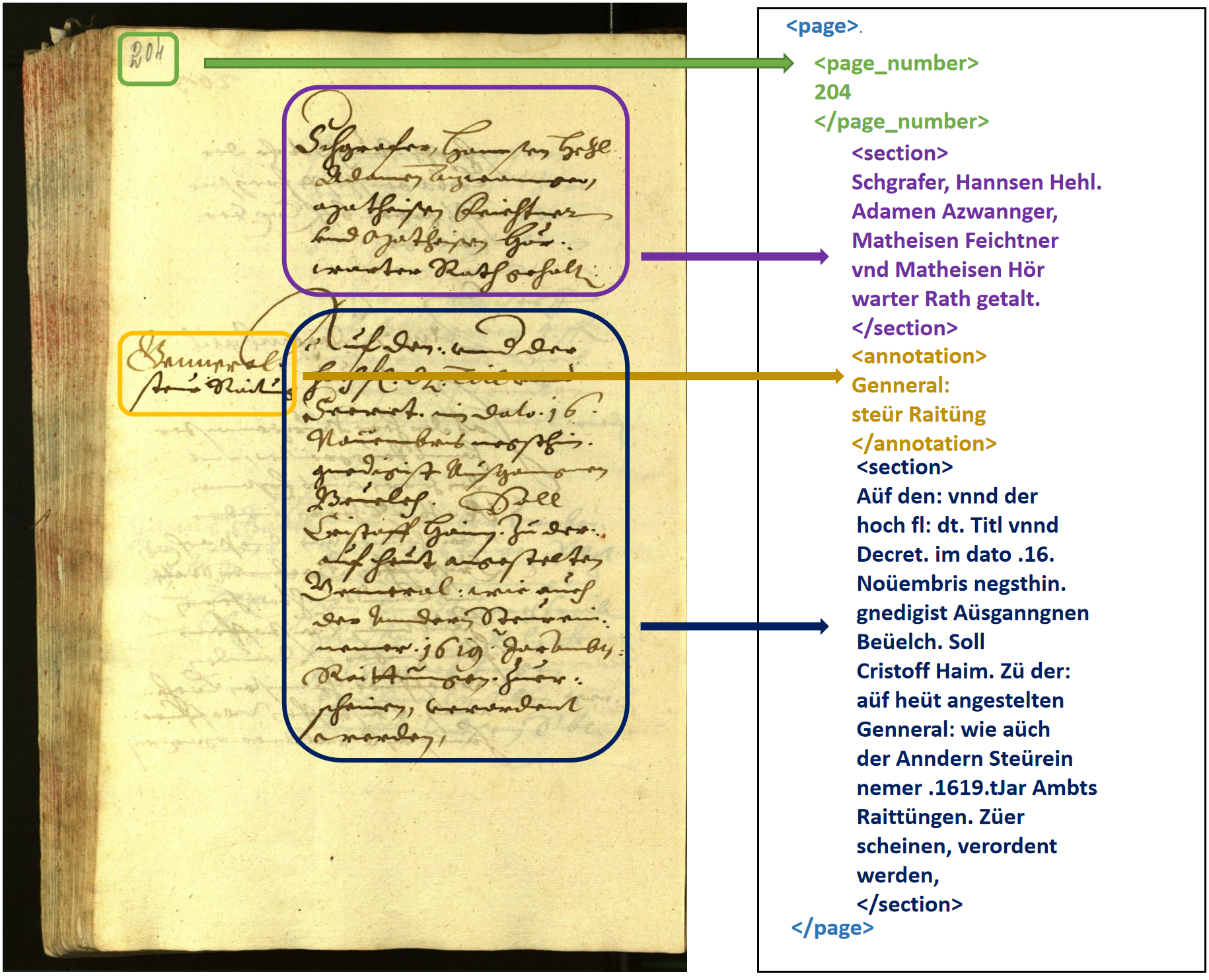}
    \vspace{-20pt}
    \caption{\textbf{Visualization of qualitative examples on the Read 2016 test set.} Our DocTTT framework predicts layout entities (shown as a graph on the right) as well as text (written in different colors between each layout entity).}
    \label{qualitative examples1}
\end{figure}
\setcounter{figure}{6}
\begin{figure}
    \centering
    \includegraphics[width=1\linewidth]{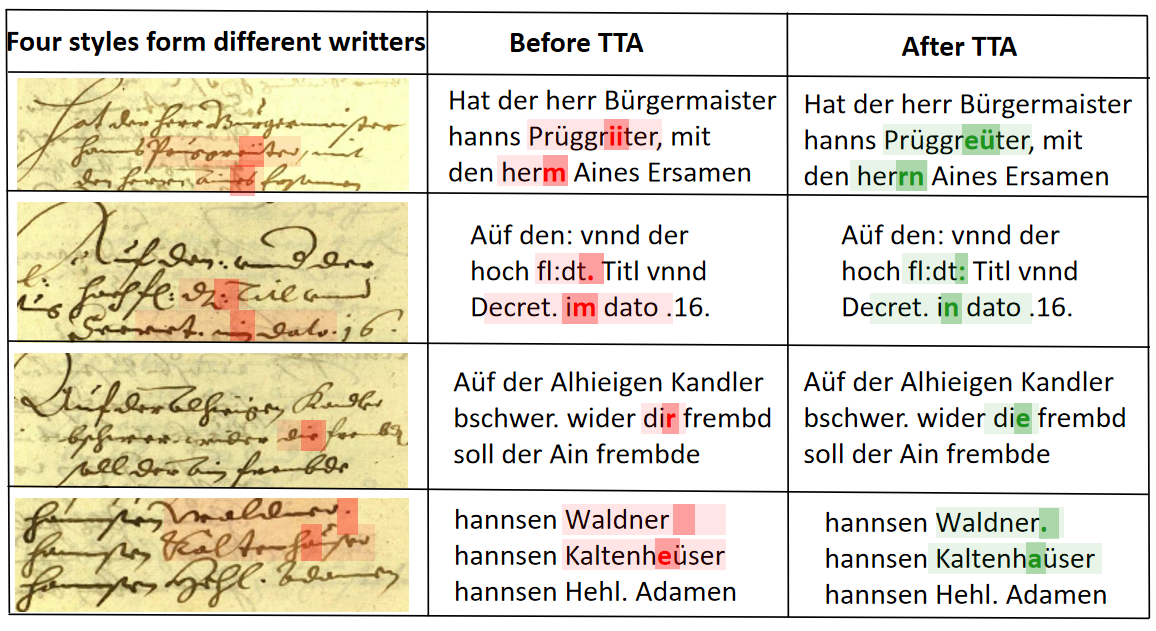}
    \vspace{-20pt}
    \caption{\textbf{Comparison of handwriting transcription corrections by DocTTT across four distinct writing styles.} This figure shows original handwriting samples (left), the initial prediction marked with errors highlighted in red (center), and the corrected prediction after applying DocTTT, highlighted in green (right). This demonstrates the effectiveness of our method in improving transcription accuracy.}
    \label{qualitative examples2}
\end{figure}

\subsection{Ablation Study}
We perform extensive ablation study to explore the impact of various architectural components on the performance of DocTTT by removing each component from DocTTT. The results are shown in Table~\ref{tab:faster_dan_ablation_study}. 
 In this table, ``Baseline'' represents the standard supervised learning approach without MAE or meta-learning, incorporating techniques like positional encoding, teacher forcing, and curriculum dropout. Removing positional encoding or teacher forcing significantly impacts performance, as positional encoding provides essential spatial context, helping the model understand the placement of text elements. Without it, the model struggles with the order and placement of characters. Teacher forcing, which assists the model in learning and self-correction, is crucial for improving prediction accuracy. Curriculum dropout also enhances performance by adding randomness to training, preventing overfitting, and promoting generalization.

For ``DocTTT w/o TTT'' in Table~\ref{tab:faster_dan_ablation_study}, we still train the model parameters in the same way as DocTTT. But we do not perform the model adaptation during testing. The results show that the meta-training procedure generally improves performance even without model adaptation during testing.

For ``DocTTT w/o meta-learning'' in Table~\ref{tab:faster_dan_ablation_study}, we employ a multi-task loss that integrates primary and auxiliary task losses, with the auxiliary task acting as a regularization mechanism during training. Post-training, we perform test-time training similarly to DocTTT, paralleling strategies from \cite{sun2020test}. The findings confirm that using MAE as an auxiliary task enhances regularization and underscores the value of model adaptation at test time.

The last row in Table~\ref{tab:faster_dan_ablation_study} is our DocTTT method that uses all components. Our method achieves the best performance in most cases. This demonstrates the importance of all components in our DocTTT framework.

\section{Conclusion}

This paper proposes a novel test-time training approach (DocTTT) for handwritten document recognition using Meta-Auxiliary learning. Instead of learning a fixed model, our method uses test-time training, so that the model parameters can be adapted to each test input. The modification is achieved via a self-supervised auxiliary task with an MAE loss. Since the auxiliary task is self-supervised, we do not need any ground-truth label for model adaptation during testing. The model parameters are learned using a meta-learning framework. The auxiliary task is learned in a way so that the adapted model can improve the performance of the primary task of HDR. Our experimental results demonstrate that DocTTT outperforms existing state-of-the-art approaches in HDR.


{\small
\bibliographystyle{ieee_fullname}
\bibliography{egbib}
}

\end{document}